\documentclass[runningheads]{llncs}
\usepackage[T1]{fontenc}
\usepackage{graphicx}
\usepackage{amsmath}
\begin{document}
\title{Advancing Cucumber Disease Detection in Agriculture
through Machine Vision and Drone Technology}
%
%
\author{Syada Tasfia Rahman
\and Nishat Vasker\
\and Amir Khabbab Ahammed
\and *Mahamudul Hasan}
%

%
\institute{Department of Computer Science and Engineering \\ East West University, Dhaka Bangladesh\\
\email{tasfiarahman70@gmail.com,nishatvasker@gmail}\\
\email{amirkhabbab99@gmail.com,munna09bd@gmail.com}\\
}
\maketitle              

\begin{abstract}
This study uses machine vision and drone technologies to propose a unique method for the diagnosis of cucumber disease in agriculture. The backbone of this research is a painstakingly curated dataset of hyperspectral photographs acquired under genuine field conditions. Unlike earlier datasets, this study included a wide variety of illness types, allowing for precise early-stage detection. The model achieves an excellent 87.5\% accuracy in distinguishing eight unique cucumber illnesses after considerable data augmentation. The incorporation of drone technology for high-resolution images improves disease evaluation. This development has enormous potential for improving crop management, lowering labor costs, and increasing agricultural productivity. This research, which automates disease detection, represents a significant step toward a more efficient and sustainable agricultural future.

\end{abstract}
\section{Introduction}
At present, manually identifying cucumber disease is lengthy and inadequate for both producers and consumers. The creation of a machine vision-based model is important to improving productivity and cutting expenses in the agricultural sector. This cutting-edge model would automate the identification of several cucumber diseases, reducing labor costs and manufacturing time. 
An exceptionally good dataset is crucial for the effective creation of the classification model. Researchers can identify and categorize cucumber diseases early on thanks to the hyperspectral images of these diseases in the given dataset ~\cite{b1}. For creating an effective agricultural automation system, this dataset ~\cite{b1} is a great resource. This dataset displays a broad range of disease classes, in contrast to previous datasets that contain scant data and concentrate on binary classification  (fresh vs. affected). More precise disease diagnosis and enhanced model performance are made possible by the increased diversity of diseases. 
Drone technology has been used to improve the disease identification process even more in addition to the usage of a machine vision-based model. Drones with high-resolution cameras and hyperspectral sensors are effective at taking pictures of cucumber fields from the air. Effective data management solutions are essential for successfully implementing this integrated approach. In order to provide real-time and precise disease evaluations, the data gathered by drones are processed, examined, and integrated into the disease detection model. Additionally, by examining patterns and trends in past drone data, researchers and practitioners can learn important insights about changes in disease prevalence and crop health over time.

\section{Related Works}
Using visible spectroscopy, ~\cite{b2} found that an online field application for cucumber disease detection is possible. Cucumber sickness detection was studied using an artificial neural network ~\cite{b5}. Loop-mediated amplification (LAMP), an isothermal approach, was used to produce a fast and sensitive Psl detection method ~\cite{b6}. The research developed a loop-mediated isothermal amplification (LAMP) technique to identify Fusarium oxysporum f. sp. cucumerinum Owen. To design a sensitive and specific pathogen identification system. LAMP's ability to accurately identify F. oxysporum f. sp. cucumerinum Owen was investigated in ~\cite{b7}. Researchers created and validated a loop-mediated isothermal amplification (LAMP) test to identify Fusarium oxysporum f. sp. cubense (FOC) with good specificity and sensitivity ~\cite{b7}. Cucumber diseases in Java were examined in ~\cite{b8}. Researchers also sought to identify and molecularly characterize the virus causing these illnesses. The authors ~\cite{b10} present a computer vision-based cucumber sickness diagnosis and categorization approach. Another influential work is ~\cite{b3}, ~\cite{b4}, ~\cite{b9}, and ~\cite{b11}. A distributed system using a wireless acoustic sensor network and machine learning techniques to detect and locate illegal drones was presented in a recent study. The technology addresses drone incursion fear with a dependable and effective solution. A comprehensive examination of the system's design and operation \cite{b13} suggests it might reduce drone dangers. Strong AI and wireless sound sensors locate illegal drones. Stats suggest this method boosts drone security in important areas. Machine learning is used to estimate drone locations. ~\cite{b13} presents a machine learning-based drone detection method in their 2018 seminal research. The viability of using two methods to evaluate truth comparison accuracy in detection, classification, and identification was suggested by a recent study ~\cite{b14}. These methodologies are tested for their ability to detect genuine positives, false positives, true negatives, and false negatives. Statistics on detection, classification, and identification included true positives, false positives, true negatives, and false negatives. Studying intervention results. Deep-learning drone inspection image analysis can predict damage. The method classifies graphical damages for analysis and decision-making. Recently developed ~\cite{b19} technique distinguishes normal and pathological blade portions. One-class SVM and deep features from a generic image dataset are used for unsupervised learning. Both components identify healthy and unhealthy blade portions. A full autopilot drone system with cloud capabilities and no human involvement is presented in ~\cite{b20}. Surveillance and anomaly detection are its main uses. Adaptable deep neural networks and computer vision models power the system. Prior research suggests that ~\cite{b20}'s information is ~\cite{b21}. An agricultural leaf disease detection and analysis system is required. We use researchers' cutting-edge drones. Modern technology should identify agricultural diseases better. Farmers decrease crop production and quality losses using leaf disease diagnostic systems. Another recognized work is ~\cite{b18}.

\section{Dataset}
Cucumbers have health benefits. Cucumber infections impact agriculture productivity and use. A dataset~\cite{b1} is provided for developing algorithms to detect cucumber diseases. Cucumber infections are classified into eight categories: \textbf{Anthracnose, Bacterial Wilt, Belly Rot, Downy Mildew, Pythium Fruit Rot, Gummy Stem Blight, Fresh leaves, and Fresh cucumber}. The dataset has 1280 unique photos (160 images per class). Data augmentation techniques were used to improve the dataset for deep learning models, resulting in 6400 augmented photos. In the following fig: \ref{fig1}, we can see the distribution of eight categories of cucumber diseases.

\begin{figure}
\includegraphics[width=0.8\textwidth]{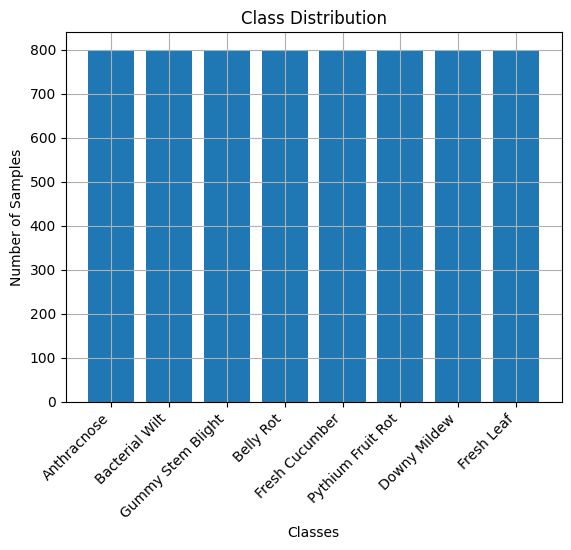}
\caption{Distribution of eight cucumber diseases.}
\label{fig1}
\end{figure}

\begin{figure}
\includegraphics[width=0.8\textwidth]{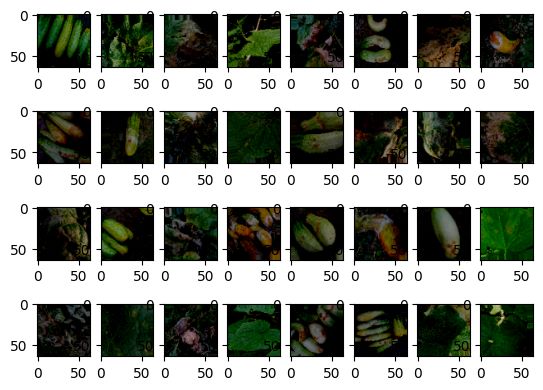}
\caption{Augmented image of different disease class} \label{fig2}
\end{figure}

In the fig: \ref{fig2}, showing the augmented image of different classes of diseases. We have resized the images into 50*50 pixels for better performance.

This dataset is a resource for the advancement of machine vision-based algorithms with drone technology for early cucumber disease diagnosis in agriculture.

\section{Materials and Methods}
\subsection{Model Description}
We have used the traditional deep learning approach VGG16 to properly train the dataset, resulting in cutting-edge findings. The validation technique entails evaluating the model's performance on the dataset in a series of phases. To guarantee that the model can acquire successfully from the dataset, data preprocessing is performed first, which includes tasks such as scaling, normalization, and augmentation. 
Training datasets teach our deep learning model. The classification model was VGG16. VGG16, 16-layer convolutional neural network. Each pixel in the input image has its expected class label in the output. VGG16 is trained on a massive class-labeled image dataset. To train VGG16, we gather hyperspectral photos of cucumber plants. The dataset covers many diseases and infectious phases. We fitted the VGG16 model to the dataset using TinyML after collection. After training, we deployed the VGG16 model to the drone. We altered the model to comply with the drone's hardware and software. We fly the drone over the cucumber field and collect real-time footage after deploying the model. The VGG16 model identifies each pixel in real-time video as healthy or unhealthy. The drone can map unhealthy cucumber fields. The illness identification system's accuracy depends on hyperspectral picture quality, training dataset size and variety, and VGG16 model hyperparameters. With an accuracy of 87.5\%, the system can identify eight distinct cucumber illnesses.

\subsection{Drone-Based Hyperspectral Imaging}
\subsubsection{Drone setup}
\begin{figure}
\includegraphics[width=0.8\textwidth]{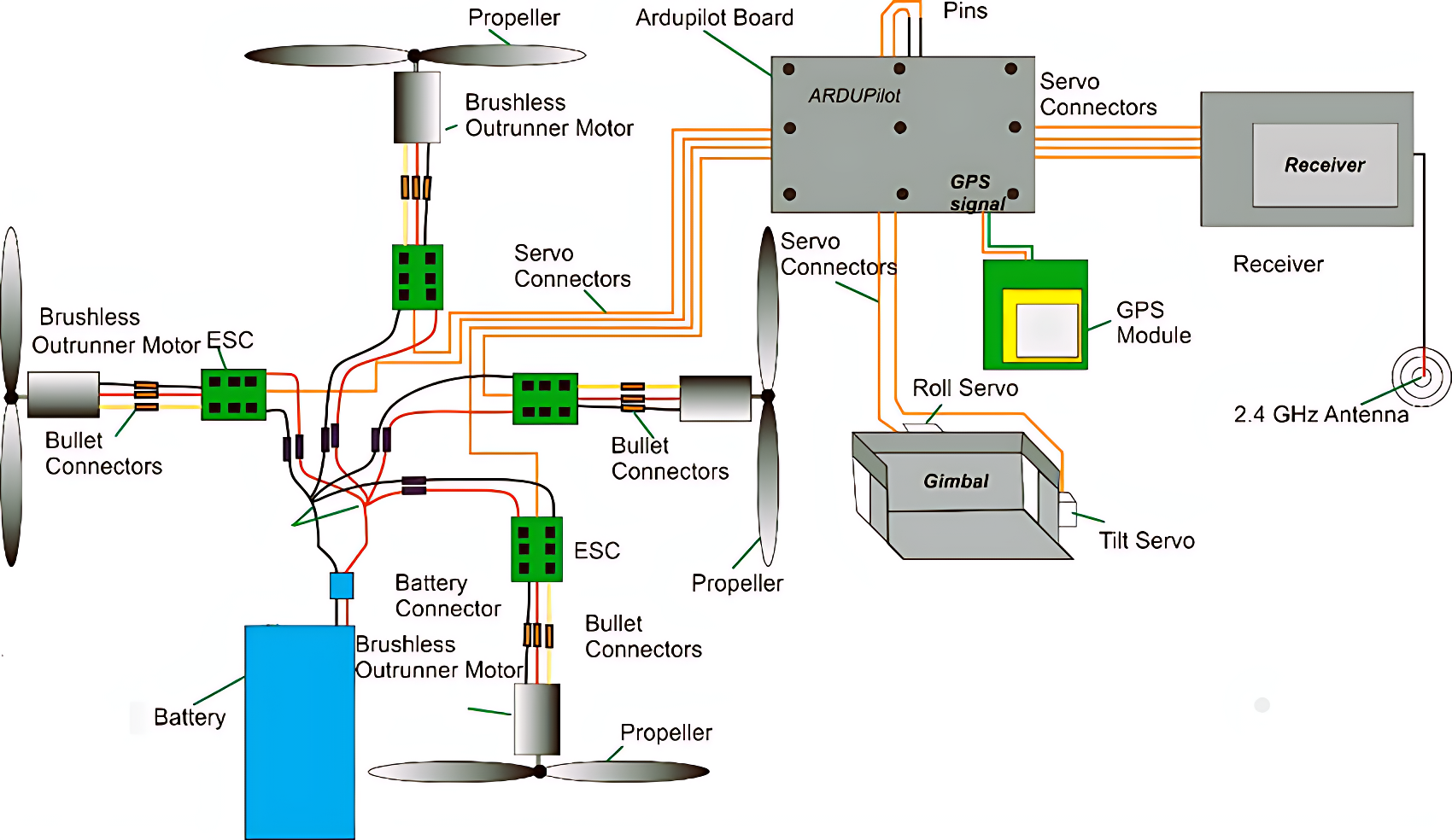}
\caption{Drone setup} \label{fig3}
\end{figure}

To facilitate the collection of hyperspectral data under genuine field conditions, we employed a custom-built drone equipped with a hyperspectral camera and real-time video streaming capabilities. The drone used in this study a (our custom drone setup diagram shown in fig: \ref{fig3}). The drone has a hyperspectral sensor that can capture photos in the visible to near-infrared spectrum.

Drones have been used to detect cucumber diseases with machine vision and TinyML by following these steps:

\begin{enumerate}
    \item Collect cucumber hyperspectral photos using a drone. The more information hyperspectral pictures have than RGB photos, the better it can identify diseases.

    \item Use TinyML to fit a VGG16 model to the hyperspectral images. TinyML is a machine learning framework that is optimized for small devices, such as drones. VGG16 is a powerful convolutional neural network that is well-suited for image classification.
    \item Deploy the VGG16 model to the drone. This has been done using a variety of software and hardware frameworks.
    \item Fly the drone over the cucumber field and capture real-time video.
    \item Feed the real-time video to the VGG16 model.
    \item The VGG16 model will classify each pixel in the video as either healthy or diseased.
    \item The drone can then be used to generate a map of the diseased areas of the cucumber field.
\end{enumerate}
This process has been summarized in the following mathematical equation:

Disease detection $=$ $f(\text{Hyperspectral images, TinyML, VGG16})$

\subsection{Data Preprocessing}

\subsubsection{Hyperspectral Data Collection}

Raw hyperspectral images were initially preprocessed to account for sensor-specific biases radiometric calibration, and atmospheric corrections. These steps were essential to ensure that the acquired data accurately represented the cucumber plants’ spectral characteristics.

\subsubsection{Hyperspectral Data Cube Generation}

The hyperspectral data were organized into data cubes, where each cube consisted of a series of 2D spectral images corresponding to different wavelengths. Mathematically, this has been represented as:

$Xi(x, y, \lambda i)= I(x, y, \lambda i)$

Where:

$ Xi(x, y, \lambda ) represents the intensity at pixel location (x,y) for a given wavelength \lambda i.$

$ I(x, y, \lambda i) represents the raw intensity measured by the hyperspectral camera.$

where:

\begin{itemize}
    \item Hyperspectral images are the images of cucumber plants that are collected by the drone.
    \item TinyML is the machine learning framework that is used to fit the VGG16 model to the hyperspectral images.
    \item VGG16 is the convolutional neural network that is used to classify each pixel in the video as either healthy or diseased.
\end{itemize}
This process has been used to detect a variety of cucumber diseases, including powdery mildew, downy mildew, and anthracnose. By detecting diseases early, farmers can take steps to protect their crops and reduce losses.

\subsubsection{Working flowchart for identification of cucumber diseases}

Drones have been used to detect cucumber diseases with machine vision and TinyML. This process is efficient and accurate, and it has the potential to improve crop management, reduce labor costs, and increase agricultural productivity.

\begin{figure}
\includegraphics[width=0.7\textwidth]{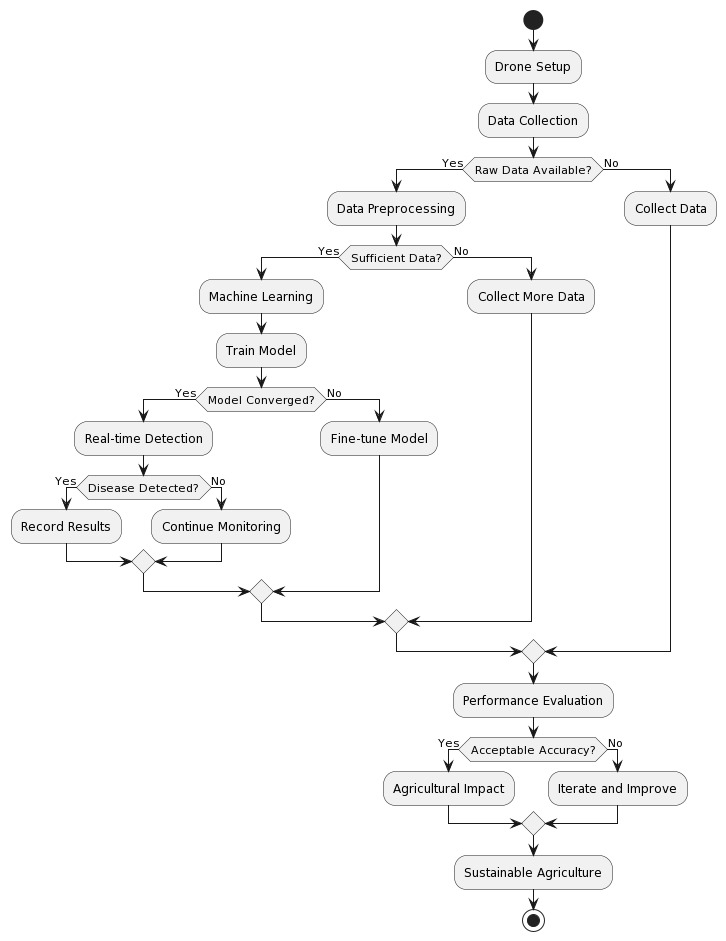}
\caption{Working flowchart} \label{fig4}
\end{figure}

\subsection{Result and Discussion}
\begin{figure}
\includegraphics[width=\textwidth]{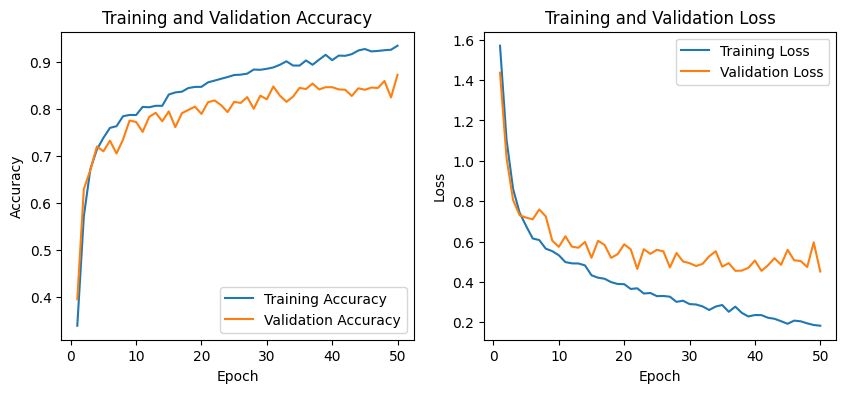}
\caption{Training and validation accuracy and loss} \label{fig5}
\end{figure}

In fig: \ref{fig5}, we can see how well the model is learning from the data during the training process and how well it generalizes unseen data during validation. It shows that our model is predicting accurately for most of the images.

Our test accuracy is 87.5\% which means it works accurately for 87.5 times within 100. From the following fig:\ref{fig6} confusion matrix shows also the accuracy of our model. It mostly predicts correctly. However, there is a little incorrect prediction in some disease classes.

\begin{figure}
\includegraphics[width=\textwidth]{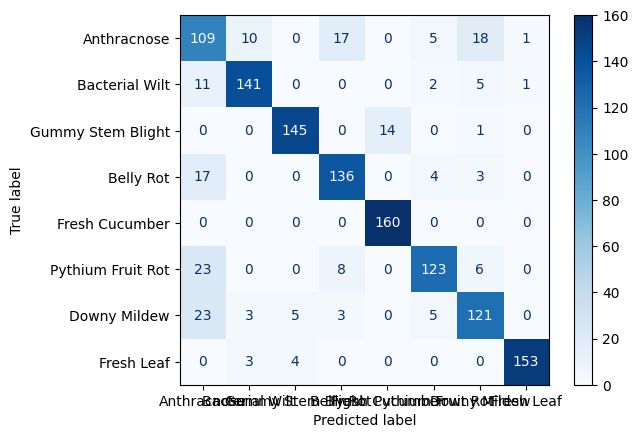}
\caption{Confusion matrix of identifying different infected classes} \label{fig6}
\end{figure}

\begin{table}
\centering

\begin{tabular}{| l |l |l |l |}
\hline
  & \textbf{Precision} & \textbf{Recall} & \textbf{f1 - Score} \\
\hline
\textbf{Anthracnose} & 0.60 & 0.68 & 0.64 \\
\hline
\textbf{Bacterial Wilt} & 0.90 & 0.88 & 0.89 \\
\hline
\textbf{Gummy Stem Blight} & 0.94 & 0.91 & 0.92 \\
\hline
\textbf{Belly Rot} & 0.83 & 0.85 & 0.84 \\
\hline
\textbf{Fresh Cucumber} & 0.92 & 1.00 & 0.96 \\
\hline
\textbf{Pythium Fruit Rot} & 0.88 & 0.77 & 0.82 \\
\hline
\textbf{Downy Mildew} & 0.79 & 0.78 & 0.77 \\
\hline
\textbf{Fresh Leaf} & 0.99 & 0.96 & 0.97 \\
\hline
\textbf{Accuracy} &  &  & 0.875 \\
\hline
\textbf{Micro Avg} & 0.86 & 0.85 & 0.875 \\
\hline
\textbf{Weighted Avg} & 0.86 & 0.85 & 0.875 \\
\hline

\end{tabular}
\caption{Classification result by predictive model} \label{table 1}
\end{table}

Table\ref{table 1} shows an evaluation of the efficacy of a classification model in recognizing various plant diseases. The evaluation includes metrics such as Precision, Recall, and F1-Score for each disease category, which collectively measure the model's performance in categorizing cases. The Micro Average and Weighted Average figures provide detailed information about the model's overall performance. The table provides a thorough summary of the classification model's efficacy in recognizing various plant diseases, with precise metrics for each category and overall performance indicators via Micro and Weighted Averages. Overall, we determined that our model's accuracy is 87.5\%, resulting in a good categorization.

\section{Conclusion}

Our research used a traditional deep learning approach to train a high-quality collection of cucumber illness photos. The validation procedure included critical processes such as data cleaning, data segmentation, model training process, and result analysis on a validation dataset before evaluating an entirely new test dataset. By using precise data pretreatment and augmentation techniques, we ensured the model's ability to train successfully from the supplied data. After training on training data, the deep learning model obtained 87.5\% accuracy on fresh testing data. The model's precision shows its ability to generalize to fresh data and forecast cucumber sickness detection and categorization. Using this machine vision-based method, farmers and agricultural specialists can improve cucumber disease diagnosis. Timely disease identification reduces crop losses and boosts agricultural output. Furthermore, this cutting-edge dataset, together with the high-performing deep learning model, provides the groundwork for future advances in automated disease identification systems, thereby helping to develop ecological and efficient crop management techniques in the agricultural sector.

%
%
%
%

\end{document}